\documentclass[11pt,a4paper]{article}
\usepackage[hyperref]{emnlp2020}
\usepackage{times}
\usepackage{latexsym}
\usepackage{graphicx} 
\usepackage{color} 
\usepackage{comment} 
\usepackage{gensymb} 
\usepackage{booktabs} 
\usepackage{colortbl} 
\usepackage{enumitem} 
\usepackage{arydshln} 

\usepackage{amsmath}
\DeclareMathOperator*{\argmin}{argmin}

\usepackage{multirow}

\usepackage{microtype}

\aclfinalcopy 
\def\aclpaperid{373} 

\title{Generating Image Descriptions\\
	via Sequential Cross-Modal Alignment Guided by Human Gaze}

\author{Ece Takmaz$^1$, Sandro Pezzelle$^1$, Lisa Beinborn$^2$, Raquel Fern\'{a}ndez$^1$\\
	$^1$Institute for Logic, Language and Computation, University of Amsterdam\\
	$^2$Vrije Universiteit Amsterdam \\
	\texttt{\{e.takmaz|s.pezzelle|raquel.fernandez\}@uva.nl},\\
	\texttt{l.m.beinborn@vu.nl}} 

\date{}

\begin{document}
\maketitle

\begin{abstract}
When speakers describe an image, they tend to look at objects before mentioning them. In this paper, we investigate such sequential cross-modal alignment by modelling the image description generation process computationally. We take as our starting point a state-of-the-art image captioning system and develop several model variants that exploit information from human gaze patterns recorded during language production. 
In particular, we propose the first approach to image description generation where visual processing is modelled \emph{sequentially}. Our experiments and analyses confirm that better descriptions can be obtained by exploiting gaze-driven attention and shed light on human cognitive processes by comparing different ways of aligning the gaze modality with language production. We find that processing gaze data sequentially leads to descriptions that are better aligned to those produced by speakers, more diverse, and more natural---particularly when gaze is encoded with a dedicated recurrent component. 
\end{abstract}


\section{Introduction}

Describing an image requires the coordination of 
different modalities.
There is a long tradition of cognitive studies showing that 
the interplay between language and vision 
is complex. On the one hand, eye movements are influenced by the task at hand, such as locating objects or verbally describing an image \cite{buswell1935people,yarbus1967eye}. On the other hand, visual information processing plays a role in guiding linguistic production \cite[e.g.,][]{griffin,gleitman2007give}. 
Such cross-modal coordination unfolds sequentially in the specific task of image description \cite{scanpatterns}---i.e., objects tend to be looked at before being mentioned.
Yet, the temporal alignment between the two modalities is not straightforward \cite{griffinbock,vaidyanathan2015alignment}

\begin{figure}[t!]\centering
	\includegraphics[width=1\columnwidth]{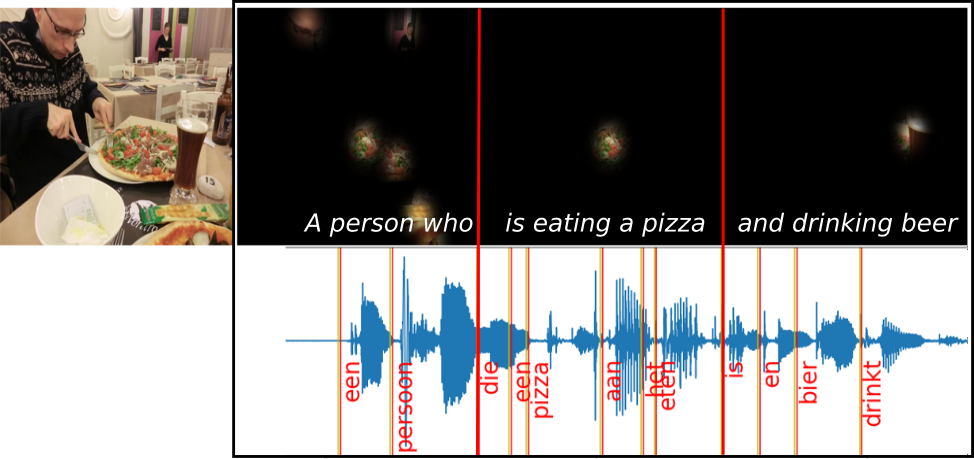}
\caption{In our approach, an image captioning model is fed with a \emph{sequence} of masked images encoding the gaze fixations of a \emph{single} human speaker during language production. This diagram is a toy illustration.}
\label{fig:example}
\end{figure}

In this paper, we follow up on these findings 
and investigate cross-modal alignment in image description by modelling the description generation process computationally. We take a state-of-the-art system for automatic image captioning \cite{DBLP:journals/corr/AndersonHBTJGZ17} and develop several model variants that exploit information derived from eye-tracking data. To train these models, we use a relatively small dataset of image descriptions in Dutch \cite[DIDEC;][]{miltenburg2018DIDEC} that includes information on gaze patterns collected during language production. 
We hypothesise that
a system that encodes gaze data as a proxy for human visual attention will lead to better, more human-like descriptions. 
In particular, we propose that training such a system with eye-movements sequentially aligned with utterances 
(see Figure~\ref{fig:example})
will produce descriptions that reflect the complex coordination 
across modalities observed in cognitive studies.

We develop a novel metric that measures the level of semantic and sequential alignment between descriptions and use it in two ways. First, we analyse 
cross-modal coordination in the DIDEC data, finding that the product of content and sequentiality better captures cross-modal correlations than content alone.
Second, we test whether our models generate captions that capture sequential alignment. Our experiments show that exploiting gaze-driven attention helps enhance image caption generation, and that processing gaze patterns sequentially results in descriptions that are better aligned with those produced by speakers, as well as being more diverse---both in terms of variability per image and overall vocabulary---particularly when gaze is encoded with a dedicated recurrent component 
that can better capture the complexity of the temporal alignment across modalities. Our data and code are publicly available at \url{https://github.com/dmg-illc/didec-seq-gen}.

Overall, this work presents the first computational model of image description generation where both visual and linguistic processing are modelled sequentially, and lends further support to cognitive theories of sequential cross-modal coordination. 


\section{Related Work}
\label{sec:related}

\paragraph{Image captioning}

Various models have been proposed to tackle the challenging task of generating a caption for a visual scene \citep{Bernardi:2016:ADG:3013558.3013571}. Contemporary approaches make use of deep neural networks and encoder-decoder architectures \cite{DBLP:journals/corr/SutskeverVL14}. In the influential model by \citet{DBLP:journals/corr/VinyalsTBE14}, a Convolutional Neural Network (CNN) is used to encode the input image into a feature representation, which is then decoded by a Long Short-Term Memory network~\cite[LSTM;][]{hochreiter1997long}
that acts as a generative language model.
In recent years, there have been many proposals to enhance this basic architecture. For instance, 
via extracting features from a lower layer of a CNN, \citet{DBLP:journals/corr/XuBKCCSZB15} obtain representations for multiple regions of an image over which attention can be applied by the LSTM decoder. 
The `Bottom-up and Top-down Attention' model by \citet{DBLP:journals/corr/AndersonHBTJGZ17} further refines this idea by extracting multiple image features with the help of Faster R-CNN \citep{NIPS2015_5638}, which results in the ability to focus on regions of different sizes better aligned with the objects in the image. Other models based on unsupervised methods \cite[e.g.,][]{8953817} and Generative Adversarial Networks \cite{DBLP:journals/corr/abs-1805-07112} have also been proposed recently.

We take as our starting point the model by \citet{DBLP:journals/corr/AndersonHBTJGZ17} for two main reasons: first, it is among the best-performing architectures on standard image captioning benchmarks; second, its underlying idea (i.e., bottom-up and top-down attention) is explicitly inspired by human visual attention mechanisms \cite{buschman2007top}, which makes it suitable for investigating the impact of adding human gaze information.

\paragraph{Eye tracking}

In computer vision, human eye movements collected with eye-tracking methods 
have been exploited to model what is salient in an image or video for object detection \cite{pascalvoc}, image classification~\cite{DBLP:journals/corr/KaressliABS16}, image segmentation \cite{Staudte2014TheIO}, region labelling \cite{vaidyanathan2015alignment,vaidyanathan2018snag}, and action detection \cite{DBLP:journals/corr/abs-1801-01582}. More relevant for the present study, gaze 
 has also been used in automatic description generation tasks, such as video frame captioning \cite{Yu2017SupervisingNA} and image captioning \cite{DBLP:journals/corr/SuganoB16,chen2018boosted,he2019human}.
In all these approaches, gaze data from different participants is 
\emph{aggregated} into a \emph{static} saliency map
to represent an abstract 
notion of saliency. This aggregated gaze data is used as supervision to train models that predict 
generic
visual saliency. 

In contrast, in our approach, we model the production process of a \emph{single} speaker by directly inputting information about where that speaker looks at during description production, and compare this to the aggregation approach. In addition, we exploit the \emph{sequential} nature of gaze patterns, i.e., the so-called
scanpath, and contrast this with the use of static saliency maps. 
Gaze scanpaths have been used in NLP 
for diverse purposes: For example, to aid part-of-speech tagging \cite{barrett-etal-2016-weakly} and chunking~\cite{klerke-plank-2019-glance}; to act as a regulariser in sequence classification tasks \cite{barrett-etal-2018-sequence}; as well as for automatic word acquisition \cite{qu2008incorporating} and
reference resolution \cite{kennington-etal-2015-incrementally}.
To our knowledge, the present study is the first attempt to investigate sequential gaze information for the specific task of image description generation.


\section{Data}
\label{sec:data}

We utilise the Dutch Image Description and Eye-Tracking Corpus \cite[DIDEC;][]{miltenburg2018DIDEC}. In particular, we use the data collected as part of the description-view task in DIDEC, where participants utter a spoken description in Dutch for each image they look at. The gaze of the participants is recorded with an
SMI RED 250
eye-tracking device
while they describe an image.
Overall, DIDEC consists of 4604 descriptions in Dutch (15 descriptions per image on average) for 307 MS COCO images \cite{lin2014coco}. For each description, the audio, textual transcription, and the corresponding eye-tracking data are provided.

\subsection{Preprocessing}

We tokenise the raw captions, lowercase them, and exclude punctuation marks and information tokens indicating, e.g., repetitions ($<$rep$>$). We then use CMUSphinx\footnote{\url{https://cmusphinx.github.io/}}
to obtain the time intervals of each word given an audio file and its transcription. See
Appendix~\ref{app:alignment} for more details. 

Gaze data in DIDEC is classified into gaze events such as fixations, saccades or blinks. 
We discard saccades and blinks (since there is no visual input during these events) and use only fixations that fall within the actual image. We treat consecutive occurrences of such fixations
as belonging to the same fixation window. 

\subsection{Saliency maps}
\label{sec:saliency}
Using the extracted fixation windows, we create two types of saliency maps, \emph{aggregated} and \emph{sequential}, which indicate the prominence of certain image regions as signalled by human gaze.

\paragraph{Aggregated saliency maps \emph{(per image)}} 
The aggregated saliency map of an image is computed as the combination 
of all participants' gazes and represents what is generally prominent given the image description task.
To create it, we first compute the saliency map of each participant who looked at the given image. 
Following \citet{integrate}, 
for each fixation window of the participant, we create a Gaussian mask 
centered at the window's centroid with a standard deviation of $1\degree$ of visual angle.
Given the data collection setup of DIDEC, this standard deviation corresponds to 44 pixels.
We 
sum up the masks weighted by relative fixation durations
and normalise the resulting mask to have values in the range $[0,1]$. Finally, we sum up and normalise the maps of all relevant participants to obtain the aggregated saliency map per image.

\paragraph{Sequential saliency maps \emph{(per image-participant pair)}} 
A sequential saliency map consists of a sequence of saliency maps aligned with the words in a description, 
and represents the scan pattern of a given participant over the course of description production. 
Using the temporal intervals extracted from the audio files, we align each word with the image regions fixated by the participant right before the word was uttered. For each word $w_t$---using the same method described above for aggregated maps---we combine all the fixation windows that took place between $w_{t-1}$ and the onset of $w_t$ and normalise them to obtain a word-level saliency map.\footnote{For the first word, we combine all the fixation windows that took place before its utterance.
Some participants may look at an image before uttering the first word to obtain its gist~\cite{oliva2006building}. However, we do not encode these differences in behaviour explicitly.}
This way, we obtain a sequence of saliency maps per participant description.

\subsection{Masked images and image features}

The saliency maps are used to keep visible only the image regions that were highly attended by participants and to mask the image areas that were never or rarely looked at (see Figure~\ref{fig:example}). 
We create each masked image by calculating the element-wise multiplication between the corresponding 2D saliency map and each RGB channel in the original image. 
We then extract image features from the masked images using ResNet-101 \cite{DBLP:journals/corr/HeZRS15} pre-trained on ImageNet \cite{imagenet_cvpr09}.
We take the output of the 2048-d average pooling layer as the image features to give as input to our models.


\section{Evaluation Measures}
\label{sec:evaluation}

We propose a novel metric to quantify the degree of both \emph{semantic} and \emph{sequential} alignment between two sentences. In our study, this metric will be leveraged in two ways: (1) to analyse cross-modal coordination in the DIDEC data (Section~\ref{sec:correlation}) and (2) to evaluate our generation models (Section~\ref{sec:exp}). 
For context, we first briefly review several existing metrics for automatic image captioning.

\paragraph{Image Captioning metrics}

Image caption generation is evaluated by assessing some kind of similarity between the generated caption and one or more reference captions (i.e., those written by human annotators). One of the most commonly used metrics is CIDEr~\cite{cider}, which (a) computes the overlapping \emph{n}-grams between the generated caption and the entire set of reference sentences for a given image, and (b) downweighs \emph{n}-grams that are frequent in the entire corpus via \emph{tf-idf} scores. Thus---regarding semantics and sequentiality---CIDEr scores can be affected by word order permutations, but not by the 
relative position of words in the entire caption
nor by the presence of different but semantically similar words.
Other metrics such as BLEU~\cite[which looks at $n$-gram precision;][]{Papineni:2002:BMA:1073083.1073135} and ROUGE-L~\cite[which considers \emph{n}-gram recall;][]{Lin:2004} suffer from comparable limitations.

METEOR~\cite{banerjee-lavie-2005-meteor} and SPICE~\cite{DBLP:journals/corr/AndersonFJG16} 
also make use of $n$-grams (or tuples in a scene's graph, in the case of SPICE) and take into account semantic similarity by matching synonyms using WordNet~\cite{pedersen2004wordnet}.
This allows for some flexibility, but can be too restrictive to grasp overall semantic similarity. To address this,
\citet{kilickaya-etal-2017-evaluating} proposed using WMD,
which builds on \texttt{word2vec} embeddings~\cite{Mikolov:2013:DRW:2999792.2999959}; more recently, several metrics capitalising on contextual embeddings~\cite{devlin2019bert} were proposed, such as BERTScore~\cite{Zhang2020BERTScore} and MoverScore~\cite{zhao2019moverscore}. However, these metrics neglect the sequential alignment of sentences.\footnote{Moreover, metrics based on contextual embeddings have been shown to suffer with languages other than English.}

\paragraph{SSD}
We propose
\emph{Semantic and Sequential Distance} (SSD),
a metric which takes into account both 
semantic similarity and the overall relative order of words.
Regarding the latter, SSD is related to \emph{Ordering-based Sequence Similarity}~\cite[OSS;][]{oss}, a measure used  by \citet{cocokeller2010} to compare sequences of categories representing gaze patterns.\footnote{Despite its name, OSS is a \emph{distance} measure. Note that it accounts for relative position, but not for semantic similarity.}
Given two sequences of words, i.e., a generated sentence $G$ and a reference sentence $R$,
SSD 
 provides a single positive value representing the overall \emph{dissimilarity} between $G$ and $R$: the closer the value to 0, the higher the similarity between the two sentences (note that the value is unbounded). 
This single value is the average of
two terms, \emph{gr} and \emph{rg}, which quantify the overall distance between $G$ and $R$---the sum of their cosine ($cos$) and positional ($pos$) distance---from $G$ to $R$ and from $R$ to $G$, respectively. The equation for \emph{gr} is given below:
\begin{equation}
gr = \sum_{i=1}^{N} cos(G\textsubscript{i},R_s(i)) + pos(G\textsubscript{i},R_s(i))
\label{eq:eq1}
\end{equation}
\noindent
where $R_s(i)$ is the semantically closest element to $G_i$ in $R$, and $cos$ in our experiments is computed over \texttt{word2vec} embeddings trained on the 4B-token corpus in Dutch, COW~\cite{tulkens2016evaluating}.

\begin{figure}[t]\centering
\includegraphics[width=\columnwidth]{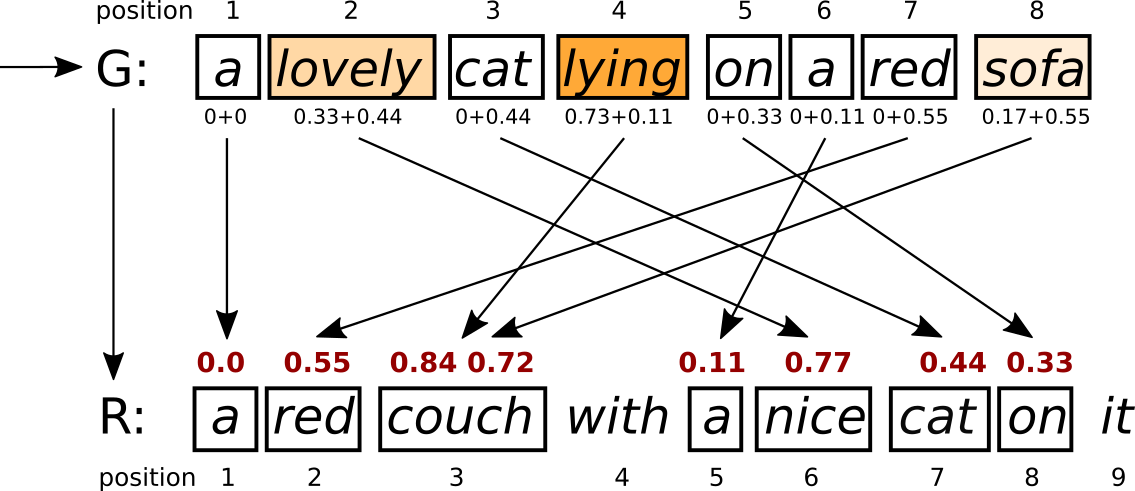}
\caption{SSD. Computation of \emph{gr} (Eq.~\ref{eq:eq1}). Sums below each word in $G$ stand for $cos+pos$, darker shades of orange for higher $cos$ distance. Value of $gr$ is the sum of numbers in red (here 3.76). 	Best viewed in color.}
\label{fig:soss}
\end{figure}

Figure~\ref{fig:soss} illustrates how the metric works in practice. Full details are in Appendix~\ref{app:ssd}. For simplicity, the diagram only shows the computation in the $gr$ direction. For example, consider the second element in $G$, `lovely'. Its closest embedding in $R$ is `nice' ($\emph{cos}=0.33$). For each of these elements, we retrieve their position index (i.e., 2 for `lovely' in $G$ and 6 for `nice' in $R$), compute their positional distance, and normalise it by the length of the longest sentence in the pair (here $R$), obtaining $|2-6|/9 \approx 0.44$. We then sum up the cosine distance and the positional distance to obtain a score for `lovely': $0.33+0.44=0.77$.
To obtain the overall \emph{gr} value, we add up the scores for all words in $G$. We compute \emph{rg} in a similar manner and obtain SSD as follows: $SSD = (gr+rg)/2$.


\section{Cross-Modal Coordination Analysis}
\label{sec:correlation}

To empirically motivate our generation models, as a preliminary experiment we investigate the level of coordination between visual attention and linguistic production in the DIDEC dataset. In particular, we test whether scanpath similarity and sentence similarity are correlated and whether taking into account the 
sequential nature of the two modalities results in
higher cross-modal alignment.

We transform gaze data into time-ordered sequences of object labels, i.e., scanpaths, (e.g., \emph{S} = `cat', `person', `cat', `table') using the annotations of object bounding boxes in the MS COCO image dataset. 
On average, scanpaths have a length of 23.4 object labels.
As for captions, we simply take the full sentences and treat them as  
sequences of words (e.g., \emph{C} = `a cute cat cuddled by a boy'). Descriptions contain an average of 12.8 tokens.

\paragraph{Order-sensitive analysis (\emph{sequential})} 
For each image, we take the set of produced descriptions and compute all pairwise similarities by using SSD (see Section~\ref{sec:evaluation}). Similarly, we take the corresponding scanpaths and compute all pairwise similarities by using OSS~\cite{oss}.
We then calculate Spearman's rank correlation (one-tailed) between the two similarity lists. This way, we obtain a correlation coefficient and \emph{p}-value for each of the 307 images in the dataset. 

\paragraph{Bag of Words analysis (\emph{BoW})} 
We compare the correlation observed in the order-sensitive analysis with a BoW approach. Here, we represent a sentence as the average of the \texttt{word2vec} embeddings of the words it contains and a scanpath as a term-frequency vector. We then perform the same correlation analysis described above.

\paragraph{Random baseline (\emph{random})} 
As a sanity check, 
using the stricter order-sensitive measures, for each image we re-compute the correlation between the two lists of similarities after randomly shuffling the 
sentences and corresponding scanpaths per image. We repeat this analysis 3 times.

\paragraph{Results} 

As shown in Table~\ref{tab:corr}, the highest level of alignment is observed in the \emph{sequential} condition, where a significant positive correlation between scanpath and sentence similarities is found for 81 images out of 307 (26\%). In BoW, the level of alignment is weaker: a positive correlation is found for 73 images (24\%), with lower maximum correlation coefficients (0.65 vs.~0.49). Substantially weaker results can be seen in the \emph{random} condition. These outcomes are in line with those obtained by~\citet{scanpatterns}
in a small dataset of 576 English sentences describing 24 images.
 
Overall, the results of the analysis indicate that the product of content and sequentiality better captures the coordination across modalities compared to content alone. Yet, the fact that positive correlations are present for only 26\% of the images suggests that coordination across modalities is (not surprisingly) more complex than what can be captured by the present pairwise similarity computation, 
confirming the intricacy of the cross-modal temporal alignment \cite{griffinbock,vaidyanathan2015alignment}.
We take this aspect into account in our proposed generation models.


\section{Models}
\label{sec:model}

The starting point for our models is the one
by \citet{DBLP:journals/corr/AndersonHBTJGZ17}.\footnote{The original implementation of this model can be found at:  \url{https://github.com/peteanderson80/bottom-up-attention}. We developed our models building on the PyTorch re-implementation of the model available at: \url{https://github.com/poojahira/image-captioning-bottom-up-top-down}.} The main aspect that distinguishes this model from other
image captioning
systems is the use of Faster R-CNN \citep{NIPS2015_5638} as image encoder, which identifies regions of the image that correspond to objects and are therefore more salient---the authors refer to this type of saliency detection as ``bottom-up attention''.
Each object region $i$ is transformed into an image feature vector $v_i$. The set of region vectors $\{v_1,\dots,v_k\}$ is utilised in two ways by two LSTM modules: 
The first LSTM takes as input the mean-pooled image feature $\overline{\textit{\textbf{v}}}$ (i.e., the mean of all salient regions) at each time step, concatenated with the two standard elements of a language model, i.e., the previous hidden state and an embedding of the latest generated word. The hidden state of this first LSTM is then used by an attention mechanism to weight the vectors in $\{v_1,\dots,v_k\}$---the authors refer to this kind of attention as ``top-down". Finally, the resulting weighted average feature vector $\hat{v}_t$ is given as input to the second LSTM module, which generates the caption one word at a time.
Note that the set of region vectors $\{v_1,\dots,v_k\}$ and the mean-pooled vector $\overline{\textit{\textbf{v}}}$ are constant over the generation of a caption, while the weights over $\{v_1,\dots,v_k\}$ and hence the weighted average feature vector $\hat{v}_t$ do change dynamically at each time step since they are influenced by the words generated so far.

\begin{table}[t]
	\resizebox{\columnwidth}{!}{
	\begin{tabular}{@{}l@{\ }ccc@{}}\toprule
		& \textit{sequential} & \multicolumn{1}{c}{\textit{BoW}} & \textit{random} \\ \midrule
		\# positively corr.  & \textbf{81}             & 73                  & 52.3 $\pm$ 5.774           \\
		\% positively corr. & \textbf{0.26}           & 0.24                & 0.17 $\pm$ 0.015         \\
		Spearman's $\rho$ (min)    & 0.15           & 0.15                & 0.15 $\pm$ 0.002         \\
		Spearman's $\rho$ (max)    & \textbf{0.65} & 0.49                & 0.50 $\pm$ 0.042         \\ \bottomrule
	\end{tabular}
}
\caption{Results of the correlation analysis: number and percentage of images with statistically significant ($p$\textless{}0.05) positive correlations and range of coefficients in the three conditions. For random, avg.~over 3 runs.} 
\label{tab:corr}
\end{table}

\begin{figure*}[ht!]\centering
	\includegraphics[width=2\columnwidth]{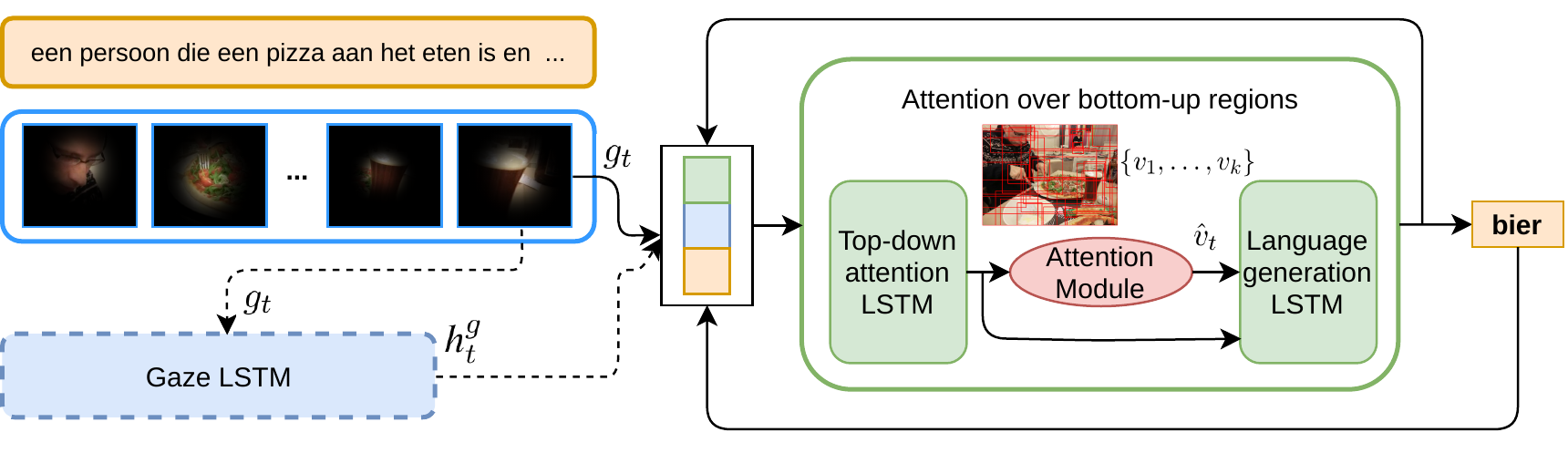}
	\caption{Architecture of the \textsc{Gaze-seq} and \textsc{Gaze-2seq} models. Dashed lines indicate that the connections to and from the Gaze LSTM are only present in the \textsc{Gaze-2seq} model.} 
	\label{fig:model_diagram}
\end{figure*}

We take the original model as our baseline and modify it to integrate visual attention defined by gaze behaviour.
In particular, we replace the mean-pooled vector $\overline{\textit{\textbf{v}}}$ by a gaze vector $g$ computed from masked images representing 
fixation patterns  
as explained in Section~\ref{sec:data}. 
We do not directly modify the set of object regions $\{v_1,\dots,v_k\}$ present in the original model (i.e., bottom-up attention is still present in our proposed models). However, the top-down attention weights learned by the models are influenced by the gaze patterns given as input. 
Concretely, we test the following model conditions: 

\begin{itemize}[leftmargin=12pt,itemsep=-1pt,topsep=2pt]
\item \textsc{no-gaze}: 
The original model as described above, with
exactly
the same image feature vectors used by \citet{DBLP:journals/corr/AndersonHBTJGZ17}.

\item \textsc{gaze-agg}: 
The mean-pooled vector $\overline{\textit{\textbf{v}}}$ in the original model is replaced with a gaze image vector $\overline{\textit{g}}$ computed on the image masked by the aggregated gaze saliency map. As explained in Section~\ref{sec:saliency}, this corresponds to the 
combination of all participants' 
fixations per image and hence remains constant over the course of generation.

\item \textsc{gaze-seq}: 
As depicted in Figure \ref{fig:model_diagram}, we replace $\overline{\textit{\textbf{v}}}$ with $g_t$, which are 
features computed for the image that was masked by the participant-specific sequential gaze saliency map at time $t$. 
Hence, $g_t$ differs at each time step $t$. Building on the results of the correlation analysis, this sequential condition thus offers a model of the production process of a speaker where visual processing and language production are time-aligned.

\item \textsc{gaze-2seq}:
Cross-modal coordination processes seem to go beyond simplistic content and temporal alignment \cite{griffinbock,vaidyanathan2015alignment}. To allow for more flexibility, we add an extra gaze-dedicated LSTM component (labelled `Gaze LSTM' in Figure \ref{fig:model_diagram}),
which processes the sequential gaze vector $g_t$ and produces a hidden representation $\textit{h}_{t}^g$. This dynamic hidden representation goes through a linear layer and then replaces $\overline{\textit{\textbf{v}}}$ at each time step $t$. 

\end{itemize}

\noindent
For the three {\sc gaze} models, we also considered a version where $\overline{\textit{\textbf{v}}}$ is concatenated with $\overline{\textit{g}}$ or $g_t$ as appropriate, rather than being replaced by the gaze vectors. Since they did not bring in better results,
we do not discuss them further in the paper.


\section{Experiments}
\label{sec:exp}

We experiment with the proposed models using the DIDEC dataset and report results per model type. 

\subsection{Setup}

We randomly split the DIDEC dataset at the image level, using 80\% of the 307 images for training, 10\% for validation, and 10\% for testing. Further details are available in Appendix~\ref{app:data}.

\begin{table*}[ht!] \centering \small
	\begin{tabular}{| l | >{\columncolor{gainsboro}}c c c  | >{\columncolor{gainsboro}}c c c |} \hline 
		\multicolumn{1}{|c|}{\it Model} & \multicolumn{3}{c|}{\it selected with SSD} & \multicolumn{3}{c|}{\it selected with CIDEr} \\ 
		&  \multicolumn{1}{c}{\bf SSD} & CIDEr & BLEU-4  &  \multicolumn{1}{c}{\bf CIDEr} & SSD & BLEU-4\\\hline
		\textsc{no-gaze}       &  5.86   (0.25)     & 55.04     (4.31)     & 39.09   (2.16)  & 52.45    (3.43)      & 6.09    (0.15)    & 35.60 (2.56)  \\ 
		\textsc{gaze-agg}     & 5.93  (0.10)      & 53.39    (3.56)      & 38.84  (1.70)   & \textbf{55.74} (3.74)     & 5.97    (0.12)    & 37.69 (1.71)   \\ 
		\textsc{gaze-seq}     & 5.82  (0.03)      & 56.16     (1.62)     & 39.80   (1.24)  & 53.59     (2.03)    & 6.10   (0.14)    & 36.09  (3.01)   \\ 
		\textsc{gaze-2seq}   & \textbf{5.81}  (0.15)      & 53.55   (1.69)       & 38.05 (1.88) & 52.94  (2.27)        & 5.93     (0.14)   & 36.27 (3.04)  \\ \hline
	\end{tabular}
	\caption{Test set results (average over 5 runs, with standard deviations in brackets) for the models selected with SSD  and with CIDEr. Scores for BLEU-4 and SSD/CIDEr when not used for model selection are shown for reference only. For SSD, lower is better; for CIDEr and BLEU-4, higher is better.}
	\label{tab:results}
\end{table*}

\paragraph{Pre-training}
Since DIDEC is a relatively small dataset, we pre-train all our models using a translated version of train/val annotations of MS COCO 2017 version. We translated all the captions in the training and validation sets of MS COCO from English to Dutch using the Google Cloud Translation API.\footnote{\url{https://cloud.google.com/translate/}} We exclude all images present in our DIDEC validation and test sets from the training set of the translated MS COCO. We randomly split the original MS COCO validation set into validation and test. The final translated 
 dataset in Dutch used for pre-training includes over 118k images for training, and 2.5k images for validation and testing, respectively, with an average of 5 captions per image.

Manual examination of a subset of translated captions showed that they are of good quality overall.
Indeed, pre-training the {\sc no-gaze} model with the translated corpus results in an improvement of about 21 CIDEr points  (from 40.81 to 61.50) in the DIDEC validation set. 
Given that the MS COCO dataset is comprised of written captions compared to DIDEC, which includes spoken descriptions, these two datasets can have distinct characteristics. We expect the transfer learning approach to help mitigate this by allowing our models to learn the features of spontaneous spoken descriptions during the fine-tuning phase.

All results reported below were obtained with pre-training (i.e., by initialising all models with the weights learned by the {\sc no-gaze} model on the translated dataset and then fine-tuning on DIDEC).

\paragraph{Vocabulary and hyperparameters}
We use a vocabulary of 21,634 tokens consisting of the union of the entire DIDEC vocabulary and the 
translated MS COCO training set vocabulary.
For all model types, we perform parameter search focusing on the learning rate, batch size, word embedding dimensions and the type of optimiser. The reported results 
refer to models trained with a learning rate of 0.0001 optimising the Cross-Entropy Loss with the Adam optimiser. The batch size is 64. The image features have 2048 dimensions and the hidden representations have 1024. The generations for the validation set were obtained through beam search with a beam width of 5. Best models were selected via either SSD or CIDEr scores on the validation set, with an early-stopping patience of 50 epochs.

More information regarding reproducibility can be found in Appendix~\ref{app:reproduce}.

\subsection{Results}
\label{sec:results}

The results obtained with different models are shown in Table~\ref{tab:results}. We report results on the test set, averaging over 5 runs with different random seeds. These scores are obtained with the best models selected on the validation set with either SSD or CIDEr.\footnote{We use the library at \url{https://github.com/Maluuba/nlg-eval} to obtain corpus-level BLEU and CIDEr scores.}
For reference, we also include scores for other metrics not used for model selection.
This allows us to check whether scores for other metrics are reasonably good when the models are optimised for a certain metric; however, only scores in the shaded columns allow us to extract conclusions on the relative performance of different model types.

On average, the best {\sc gaze} models outperform the 
{\sc no-gaze} model: 5.81 vs.\ 5.86 for SSD (lower is better)
and 55.74 vs.\ 52.45 for CIDEr (higher is better). This indicates that eye-tracking data encodes patterns of attention that can contribute to the enhancement of image description generation. 
Zooming into the different gaze-injected conditions, we find that among the models selected with SSD, the sequential models perform better than {\sc gaze-agg} 
(5.81 and 5.82 vs.\ 5.93).
This shows that the proposed models succeed (to some extent) in capturing the \emph{sequential} alignment across modalities, and that such alignment can be exploited for description generation. 
Interestingly, {\sc gaze-2seq} is the best-performing gaze model: it has the best average SSD across runs and the best absolute single run (5.70 vs.\ 5.79 and 5.80 by {\sc gaze-seq} and {\sc gaze-agg}, respectively).
This suggests that the higher flexibility and abstraction provided by the gaze-dedicated LSTM
component 
offers a more adequate model of the intricate ways in which the two modalities are aligned.

As for the CIDEr-selected models, on average the gaze-injected models also perform better than {\sc no-gaze}.
The best results are obtained with {\sc gaze-agg} (55.74).
This is consistent with what CIDEr captures: 
it takes into account regularities across different descriptions of a given image; therefore, using a saliency map that combines the gaze patterns of several participants leads to higher scores than inputting sequential saliency maps, which model the path of fixations of each speaker independently. 
This variability seems to have a negative effect on CIDEr scores of sequential models, which are lower than
{\sc gaze-agg}; yet higher than {\sc no-gaze} (53.59 and 52.94 vs.\ 52.45).

\begin{figure*}[ht!]
	\resizebox{\textwidth}{!}{
		\begin{tabular}{lllll}
			& \includegraphics[width=0.25\textwidth]{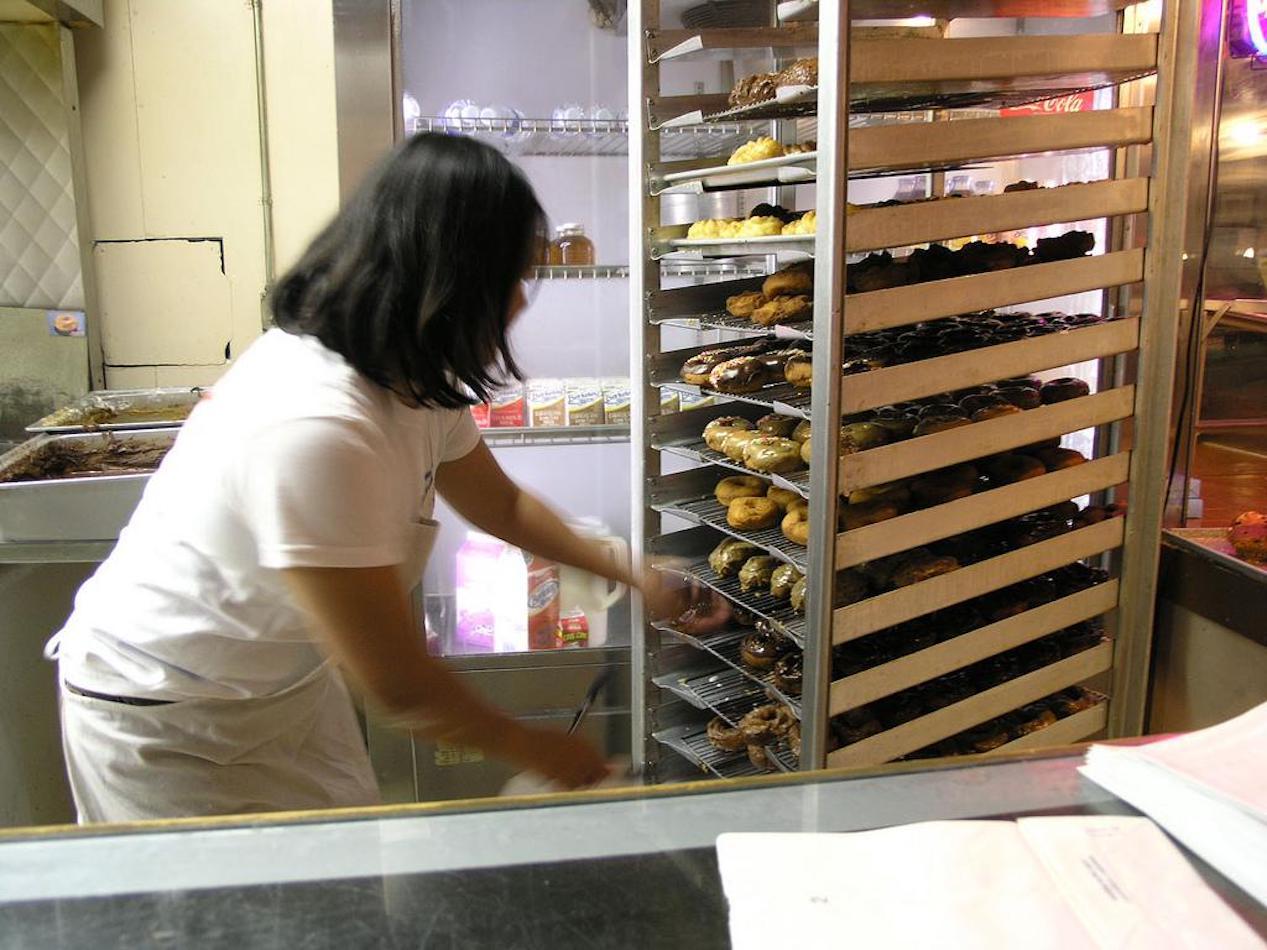} &
			\includegraphics[width=0.25\textwidth]{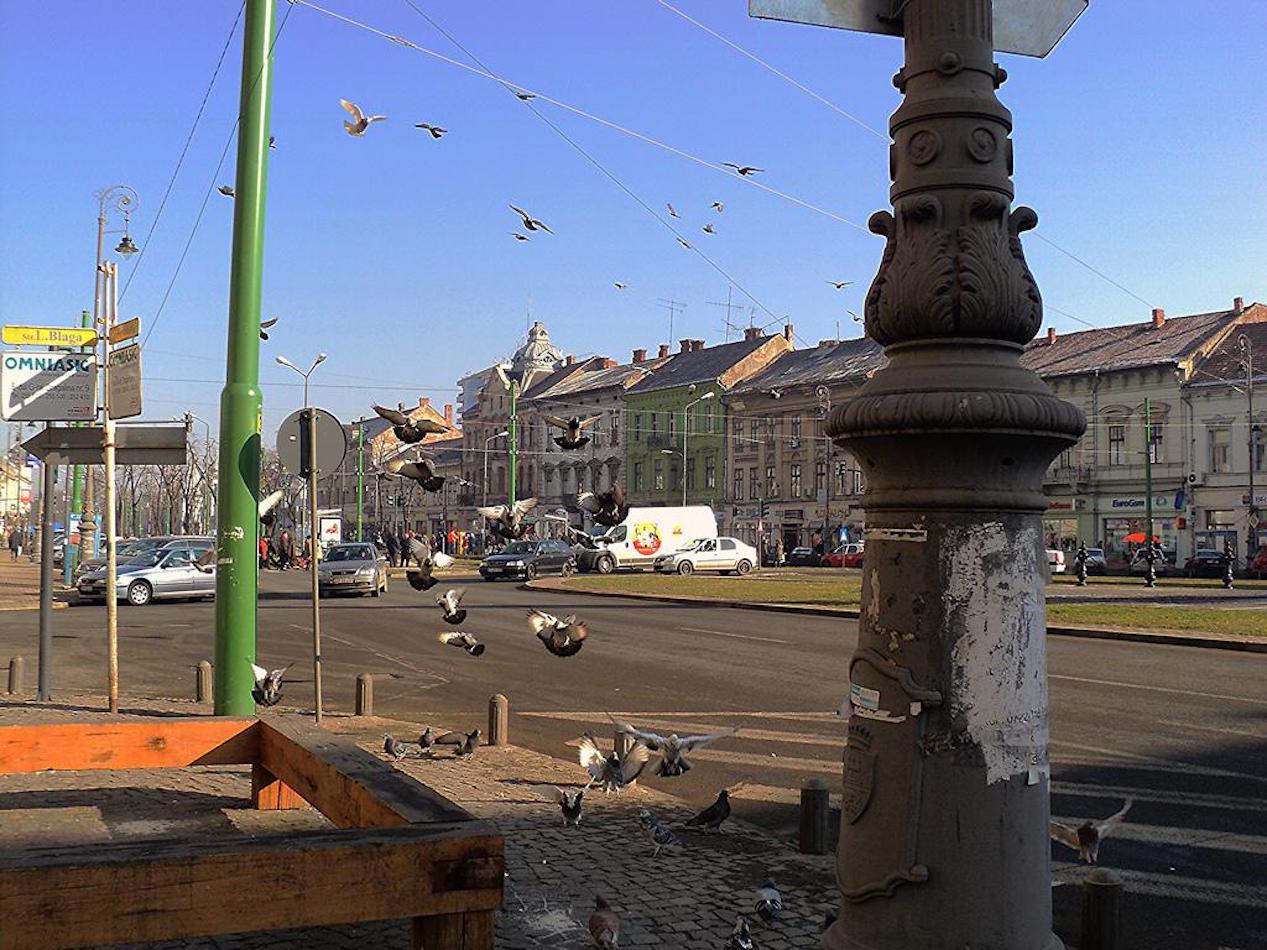}&
			\includegraphics[width=0.25\textwidth]{images/713684_two.jpg}&
			\includegraphics[width=0.25\textwidth]{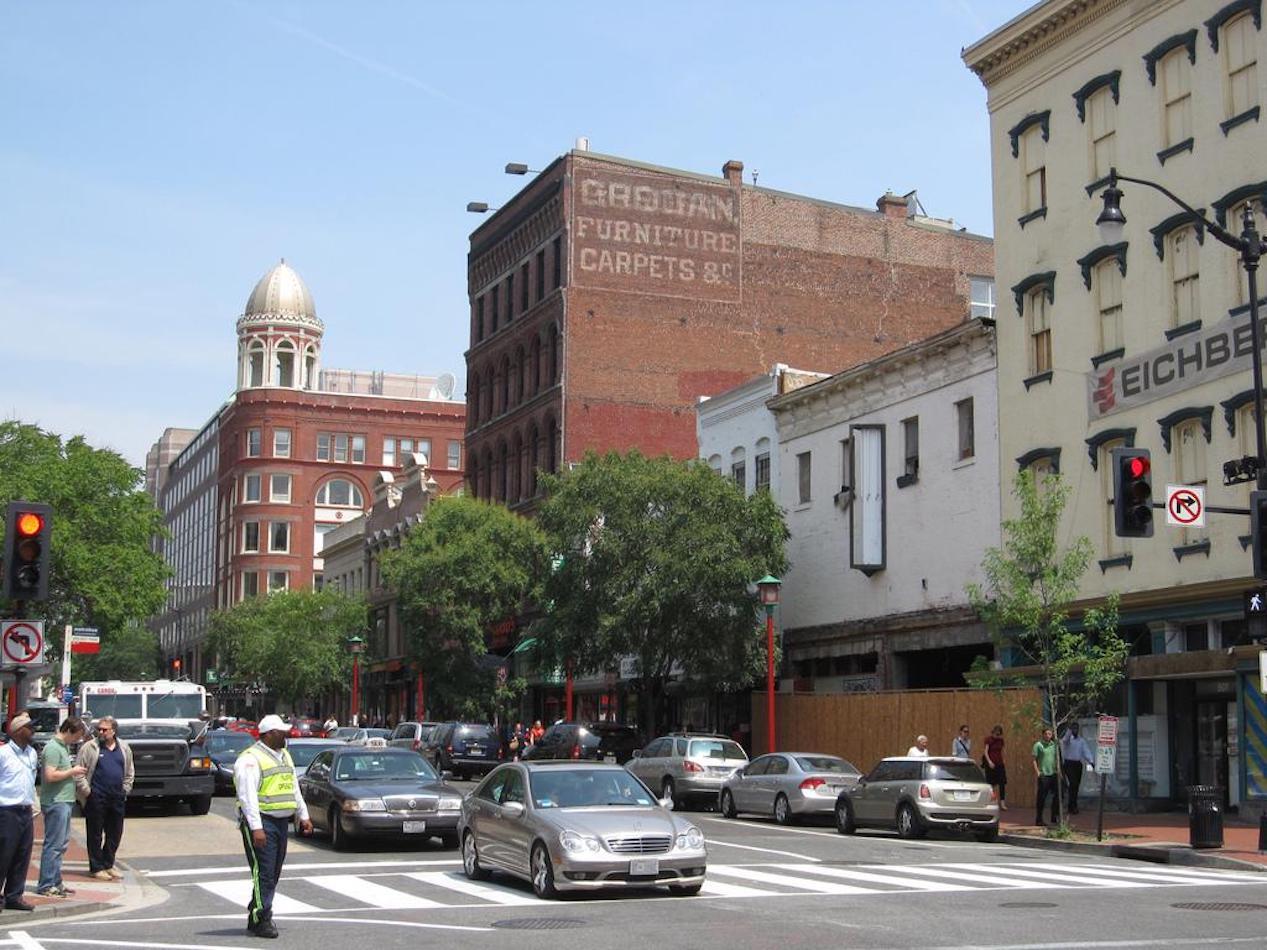}
			\vspace{0.2cm}\\

			&
			\vspace{0.2cm}
			\textbf{specificity}&\textbf{disfluency}&\textbf{compression}&\textbf{repetition}\\

			\textsc{no-g}&
			een vrouw die in de keuken staat\dots&een foto van een straat met een aantal vogels&een rode bus en een bus&
			een straat met \textbf{auto's} en \textbf{auto's}\\ 
			\vspace{0.2cm}
			&
			\emph{(a woman who is standing in the kitchen\dots)}&\emph{(a photo of a street with a number of birds)}&\emph{(a red bus and a bus)}&
			\emph{(a street with \textbf{cars} and \textbf{cars})}\\
			
			\textsc{2seq}&
			een vrouw in een keuken met \textbf{donuts}&\textbf{uh uh uh uh} met een aantal vogels&\textbf{twee} bussen die geparkeerd staan&
			een straat in de stad met \textbf{auto's} en \textbf{auto's}\\
			&
			\emph{(a woman in the kitchen with \textbf{donuts})}&\emph{(\textbf{uh uh uh uh} with some birds)}&\emph{(\textbf{two} buses that are parked)}&
			\emph{(a street in the city with \textbf{cars} and \textbf{cars})}\\

		\end{tabular}
	}
	\caption{Phenomena that are either particular to gaze models (specificity, disfluency, and compression) or  common to all (repetition). Abbreviations \textsc{no-g} and \textsc{2seq} refer to \textsc{no-gaze} and \textsc{gaze-2seq}, respectively.
	}	
	\label{fig:examples}
	
\end{figure*}

It is worth noting that CIDEr and BLEU-4 scores obtained with the SSD-selected models are sensible, which indicates that the generated descriptions do not suffer with respect to distinct aspects evaluated by other metrics  when the models are optimised with SSD.
Indeed, the highest CIDEr score obtained among models selected via SSD ({\sc gaze-seq}: 56.16) is even higher than that obtained by the best CIDEr-selected one ({\sc gaze-agg}: 55.74).
However, this is likely due to CIDEr being sensitive to lexical differences between the test set and the validation set used for model selection, which could lead to slightly different patterns.


\section{Analysis}
\label{sec:analysis}

This section presents an analysis of the descriptions generated by the models on the test set (446 descriptions). We focus on one single run per model.

\paragraph{Cross-modal sequential alignment}

Given what SSD captures, our results indicate that the captions generated by \textsc{gaze-2seq} are better aligned---in terms of semantic content and order of words---with the human captions than
the ones generated by non-sequential models. Arguably, this enhanced alignment is driven by the specific information provided by the scanpath of each speaker. If this information is used effectively by the sequential models, then we should see more variation in their output. By definition, the non-sequential models generate only one single caption per image. Are the sequential models able to exploit the variation stemming from the speaker-specific scanpaths?
{Indeed, we find that \textsc{gaze-2seq} generates an average of 4.4 different descriptions per image (i.e., 30\% of the generated captions per image  are unique).

Furthermore, we conjecture that tighter coordination between scanpaths and corresponding descriptions should give rise to more variation, since presumably the scanpath has a stronger causal effect on the description in such cases. To test this, we take the 30 images in the test set and divide them into two groups: (A) images for which a significant positive correlation was found in the cross-modal coordination analysis of Section~\ref{sec:correlation}; (B) all the others. These groups include, respectively, 10 and 20 images. As hypothesised, we observe a higher percentage of unique captions per image in A (35\%) compared to B (27\%). 

\paragraph{Quantitative analysis}
We explore whether there are any quantitative differences across models regarding two aspects, i.e., the average length in tokens of the captions, and the size of the vocabulary produced. No striking differences are observed regarding caption length: \textsc{no-gaze}  produces slightly shorter captions (avg.\ 7.5) compared to both \textsc{gaze-2seq} (avg.\ 7.7) and \textsc{gaze-agg} (avg.\ 8.1). The difference, however, is negligible. Indeed, it appears that equipping models with gaze data does not make sentence length substantially closer to the length of reference captions (avg.\ 12.3 tokens). 

In contrast, there are more pronounced differences regarding vocabulary. 
While \textsc{gaze-agg} has a similar vocabulary size (68 unique tokens produced) to \textsc{no-gaze} (63), \textsc{gaze-2seq} is found to almost double it,
with 109 unique tokens produced. Though this number is still far from the total size of the reference vocabulary (813), this trend suggests that a more diverse and perhaps `targeted'  language is encouraged when specific image regions are identified through gaze-based attention. 
The following qualitative analysis sheds some light on this hypothesis.

\paragraph{Qualitative analysis} 
Manual inspection of all
the captions generated by the models 
reveals interesting qualitative differences.
First, captions generated by gaze-injected models are more likely to refer to objects---even when they are 
small and/or in the background---which are image-specific and thus very relevant for the caption. For example, when describing the leftmost image in Fig.~\ref{fig:examples}, \textsc{no-gaze} does not mention the word \emph{donuts}, which is 
produced by both \textsc{gaze-agg} and \textsc{gaze-2seq}. Second, gaze-injected models produce language that seems to reflect uncertainty present in the visual input. For the second image of Fig.~\ref{fig:examples}, e.g., both \textsc{gaze-agg} and \textsc{gaze-2seq} generate disfluencies such as \emph{uh}
(interestingly, several participants' descriptions include similar disfluencies for this same image, which suggests some degree of uncertainty at the visual level); in contrast, in the entire test set no disfluencies are produced by \textsc{no-gaze}. 

Finally, we find that \textsc{gaze-2seq} is able to produce captions that somehow `compress' a repetitive sequence (e.g., \emph{a red bus and a bus}) into a shorter one, embedding a number
(e.g., \emph{two buses that are parked}; see third example in Fig.~\ref{fig:examples}).
This phenomenon is never observed in the output of other models (crucially, not even in \textsc{gaze-seq}). We thus conjecture that this ability is due to the presence of the gaze-dedicated LSTM, which allows for a more abstract processing of the visual input. However, the presence of gaze data does not fully solve the issue of words being repeated within the same caption, as illustrated by the rightmost example in Fig.~\ref{fig:examples}. Indeed, this weakness is common to all models, including the best performing \textsc{gaze-2seq}.


\section{Conclusions}
\label{sec:conclusions}

We tackled the problem of automatically generating an image description from a novel perspective, by modelling the sequential visual processing of a speaker concurrently with language production. Our study shows that better descriptions---i.e., more aligned with speakers' productions in terms of content and order of words---can be obtained by equipping models with human gaze data. Moreover, this trend is more pronounced when gaze data is fed \emph{sequentially}, in line with cognitive theories of sequential cross-modal alignment \cite[e.g.,][]{scanpatterns}.

Our study was conducted using the Dutch language dataset DIDEC \cite{miltenburg2018DIDEC}, which posed the additional challenges of dealing with a small amount of data and a low resource language. 
We believe, however, that there is value in conducting research with 
languages other than English. 
In the future, our approach  and new evaluation measure could be applied to 
larger eye-tracking datasets, such as the English dataset by~\citet{he2019human}.
Since different eye-tracking datasets tend to make use of different gaze encodings and formats, 
the amount of pre-processing and analysis steps required to apply our method to
other resources was beyond the scope of this paper. We leave testing whether the reported
pattern of results holds across different languages to future work.

Despite the challenges mentioned above, our experiments show that a state-of-art image captioning model can be effectively extended to encode cognitive information present in human gaze behaviour.
Comparing different ways of aligning the gaze modality with language production, as we have done in the present work, can shed light on how these processes unfold in human cognition. This type of computational modelling could help, for example, study the interaction between gaze and the production of filler words and repetitions, which we have not investigated in detail. 
Taken together, our results open the door to further work in this direction and support the case for computational approaches leveraging cognitive data.

\section*{Acknowledgments}
We are grateful to Lieke Gelderloos for her help with the Dutch transcriptions, and to Jelle Zuidema and the participants of EurNLP 2019 for their feedback on a preliminary version of the work. 
Lisa Beinborn worked on the project mostly when being employed at the University of Amsterdam. This project has received funding from the European Research Council (ERC) under the European Union's Horizon 2020 research and innovation programme (grant agreement No.~819455 awarded to Raquel Fern{\'a}ndez).

\bibliography{emnlp2020} 
\bibliographystyle{acl_natbib}

\appendix

\section*{Appendices}


\section{Audio-Caption Alignment}
\label{app:alignment}

In this appendix, we provide details on the pipeline used to time-align audio and descriptions. After processing a transcribed caption, we insert it as a grammar rule into a Java Speech Grammar Format (JSGF) file to be fed into CMUSphinx. As CMUSphinx supports English by default, we incorporated into the tool the phonetic and language models and the dictionary for Dutch as provided by the developers of CMUSphinx.\footnote{\url{https://sourceforge.net/projects/cmusphinx/files/Acoustic\%20and\%20Language\%20Models/Dutch/}} 

Some words in our JSGF files were not in the VoxForge Dutch phonetic dictionary of CMUSphinx, which lists lexical items and their corresponding pronunciations in a format similar to ARPABET, adapted for Dutch.\footnote{\url{http://www.speech.cs.cmu.edu/cgi-bin/cmudict}}
To overcome this problem, we used eSpeak\footnote{\url{http://espeak.sourceforge.net/}} to obtain the International Phonetic Alphabet (IPA) transcriptions of such out-of-vocabulary words. We obtained the set of IPA symbols existing in the transcriptions of out-of-vocabulary words and the set of ARPABET symbols in the dictionary. Then, a native speaker of Dutch, who is also a linguist, manually produced a mapping from these IPA symbols to ARPABET symbols of Dutch phonemes.\footnote{The mapping from IPA symbols to ARPABET symbols is provided in our GitHub repository.} Given this mapping, we automatically converted out-of-vocabulary tokens into the required format and appended to the dictionary. A similar approach was also followed for numbers in numeric notation and certain English words.

\begin{figure}[t!]\centering
	\includegraphics[scale=0.59]{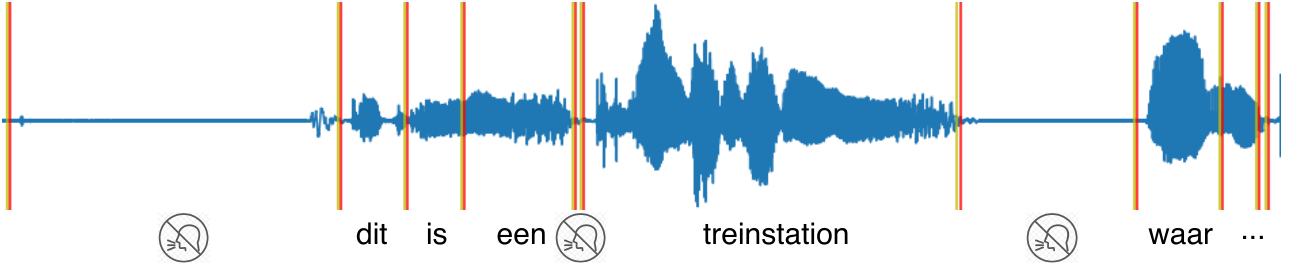}
	\caption{Temporal alignment of words in a transcribed caption and the corresponding audio file.}
	\label{fig:alg_plot}
\end{figure}

For some audio-caption pairs, the tool could not find an alignment matching the grammar. We turned off noise- and silence-removal and experimented with parameters related to beam-decoding in CMUSphinx to allow for a maximal number of complete alignments. However, we had to exclude some captions where there were unintelligible words in particular at the beginning or in the middle of the audio, since such an issue disrupts the alignment procedure.

Considering possible inter-participant differences in terms of pronunciation,  
the quality of audio files, 
and possible noise in the background of recordings, we assume that the time intervals of the words we obtained after these pre-processing steps are approximate indicators. Although there might be a few cases where the alignment is not quite accurate, we find this way of obtaining utterance timestamps reliable in general. An example audio-caption alignment is shown 
in Figure \ref{fig:alg_plot}.

\section{SSD: Further Details}
\label{app:ssd}

\begin{table*}[ht!] \centering\small
	\begin{tabular}{clccc}\toprule
		\textbf{Type} & \textbf{Description}                                                                                    & \textbf{SSD}          & $gr$          & $rg$ \\ \midrule
		$R$           & een dubbeldekker bus                                                                                    & \multirow{2}{*}{1.41} & \multirow{2}{*}{2.82} & \multirow{2}{*}{0.00}            \\ 
		$G$           & een dubbeldekker bus in een stad                                                                        &                       &                       &                                  \\ \midrule
		$R$           & een dubbeldekkerbus in uh in engeland                                                                   & \multirow{2}{*}{2.64} & \multirow{2}{*}{2.72} & \multirow{2}{*}{2.55}            \\ 
		$G$           & een dubbeldekker bus in een stad                                                                        &                       &                       &                                  \\ \midrule
		$R$           & een rustige straat met een bus tegemoetkomend naar \textless{}unk\textgreater \ nummer 43                 & \multirow{2}{*}{5.87} & \multirow{2}{*}{4.31} & \multirow{2}{*}{7.43}            \\ 
		$G$           & een dubbeldekker bus die op een weg rijdt                                                               &                       &                       &                                  \\ \midrule
		$R$          & een bus met lijn 43 die aan het rijden is waarvan uh de bus uit twee \textless{}unk\textgreater \ bestaat & \multirow{2}{*}{8.62} & \multirow{2}{*}{0.43} & \multirow{2}{*}{16.81}           \\ 
		$G$           & een dubbeldekker bus                                                                                    &                       &                       &                                  \\ \bottomrule
	\end{tabular}
	\caption{Examples of SSD scores for several descriptions generated ($G$) by \textsc{gaze-2seq}  compared to the reference description ($R$). 
		$gr$  and $rg$ indicate the direction of the calculation. Lower SSD scores are better.}
	\label{tab:examples}
\end{table*}

SSD is the average of two terms, \emph{gr} and \emph{rg}, which quantify the overall distance between a generated sentence ($G$) and a reference sentence ($R$). Eq.~\ref{eq2} (identical to Eq.~\ref{eq:eq1} in Section~\ref{sec:evaluation}) shows the calculation from $G$ to $R$ and Eq.~\ref{eq3} from $R$ to $G$:
\begin{equation}
gr = \sum_{i=1}^{N} cos(G\textsubscript{i},R_s(i)) + pos(G\textsubscript{i},R_s(i))
\label{eq2}
\end{equation}
\begin{equation}
rg = \sum_{j=1}^{M} cos(R\textsubscript{j},G_s(j)) + pos(R\textsubscript{j},G_s(j))
\label{eq3}
\end{equation}

\noindent
$N$ and $M$ refer to number of tokens in $G$ and $R$, respectively. 
Cosine and positional distances are computed between the \emph{i}\textsubscript{th} element of $G$ and another token, which is the most semantically similar word to $G_i$ in R. $R_s(i)$ is the most semantically similar word to $G_i$ and $G_s(j)$ is the most semantically similar word to $R_j$:
\begin{equation}
R_s(i) = \argmin_j(cos(G\textsubscript{i},R\textsubscript{j})) 
\end{equation}
\begin{equation}
G_s(j) = \argmin_i(cos(R\textsubscript{j},G\textsubscript{i}))
\end{equation}

\noindent 
Table \ref{tab:examples} shows some example descriptions generated by the \textsc{gaze-2seq} model and corresponding references for a single image. We report the overall SSD scores along with $gr$ and $rg$ values separately.

\section{Data Split Statistics}
\label{app:data}

Table \ref{tab:counts} lists the number of images belonging to each split after we divide the DIDEC corpus (description-view partition) with respect to the images. In addition, the total number of captions in each split is provided. 

\begin{table}[ht!] \centering \small
	\begin{tabular}{lcccc}\toprule
		\textbf{} & \textbf{train} & \textbf{val} & \textbf{test} & \textbf{total} \\ \midrule
		Images          & 247            & 30           & 30            & 307            \\ 
		Captions        & 3658           & 444          & 446           & 4548           \\ \bottomrule
	\end{tabular}
	\caption{Number of images and captions.}
	\label{tab:counts}
\end{table}

\noindent The number of human descriptions per image varies in DIDEC and as we also removed some captions during preprocessing, images do not have an equal number of captions. Therefore, we report the average number of captions per image for each split, as well as their range, in Table \ref{tab:avg_caps}.

\begin{table}[ht!] \centering \small
	\begin{tabular}{lcccc}\toprule
		 & \textbf{train} & \textbf{val} & \textbf{test} & \textbf{overall} \\ \midrule
         Avg   & 14.81 & 14.80 & 14.87 & 14.81\\
         Min   & 11 & 12 & 13 & 11 \\
         Max  & 16 & 16 & 16 & 16\\\bottomrule
	\end{tabular}
	\caption{Number of captions per image.}
	\label{tab:avg_caps}
\end{table}

\section{Reproducibility}
\label{app:reproduce}
We implemented and trained our models in Python version 3.6\footnote{\url{https://www.python.org/downloads/release/python-360/}} and PyTorch version 0.4.1.\footnote{\url{https://pytorch.org/}} All models were run on a computer cluster with Debian Linux OS. Each model used a single GPU GeForce 1080Ti, 11GB GDDR5X, with NVIDIA driver version: 418.56 and CUDA version: 10.1. 

Pre-training with the translated MS COCO dataset took approximately 5 days. \textsc{No-Gaze} and \textsc{Gaze-Agg} took around 1.5 hours and \textsc{Gaze-Seq} and \textsc{Gaze-2Seq} models took 2 hours to fine-tune over the pre-trained model.

Since the pre-trained model and the fine-tuned \textsc{No-Gaze}, \textsc{Gaze-Agg} and \textsc{Gaze-Seq} models use essentially the same architecture, they have an equal number of parameters: ~85 million. 
\textsc{Gaze-2Seq} has more parameters due to the addition of the Gaze LSTM: ~100 million. 

In all the models, the biases in linear layers were set to 0 and the weights were uniformly sampled from the range (-0.1, 0.1). Embedding weights were initialised uniformly in the range (-0.1, 0.1). LSTM hidden states were initialised to 0.

Below we give details regarding the manually-tuned hyperparameters.

\subsection{Hyperparameters for Pre-Training}
We experimented with learning rate (0.001, 0.0001), dimensions for the word embeddings and hidden representations (512, 1024) and batch size (64, 128). The best pre-trained model is selected based on its CIDEr score on the validation split of our translated MS COCO dataset, with an early-stopping patience of 20 epochs. We use a learning rate of 0.0001 optimising the Cross-Entropy Loss with the Adam optimiser. The batch size is 128. The image features have 2048 dimensions and the hidden representations 1024. The generations for the validation set are obtained through beam search with a beam width of 5. 

\subsection{Hyperparameters for Fine-tuning}
We experimented with the same set of hyperparameters as in pre-training.
The details of the hyperparameters for the selected models were given in the main text. We select the models separately based on CIDEr scores and SSD scores. We train each model type with their selected configuration with 5 different random seeds to set the random behaviour of PyTorch and NumPy. We also turn off the cuDNN benchmark and also set cuDNN to deterministic.


\end{document}